\documentclass[5p,times,procedia]{elsarticle}
\usepackage{ecrc}

\volume{00}

\firstpage{1}

\journalname{Procedia CIRP}

\runauth{M. Goebels et al.}

\jid{trpro}

\usepackage{amssymb}

\usepackage[figuresright]{rotating}

\usepackage[bookmarks=false]{hyperref}
    \hypersetup{colorlinks,
      linkcolor=blue,
      citecolor=blue,
      urlcolor=blue}

\begin{document}
\begin{frontmatter}



\dochead{57th CIRP Conference on Manufacturing Systems 2024 (CMS 2024)}%

\title{Milling using two mechatronically coupled robots}


\author[a]{Max Goebels\corref{cor1} \fnref{fn1}}
\author[a]{Jan Baumgärtner\fnref{fn1}}
\author[a]{Tobias Fuchs\fnref{fn1}}
\author[a]{Edgar Mühlbeier\fnref{fn1}}
\author[a]{Alexander Puchta}
\author[a]{Jürgen Fleischer}

\fntext[fn1]{These authors contributed equally to this work.}
\cortext[cor1]{Corresponding author. Mail: max.goebels@kit.edu}
\address[a]{wbk Institute of Production Science, Karlsruhe Institute of Technology, Karlsruhe, Germany}

\begin{abstract}
  Industrial robots are commonly used in various industries due to their flexibility.
  However, their adoption for machining tasks is minimal because of the low dynamic stiffness characteristic of serial kinematic chains.
  To overcome this problem, we propose coupling two industrial robots at the flanges to form a parallel kinematic machining system.
  Although parallel kinematic chains are inherently stiffer, one possible disadvantage of the proposed system is that it is heavily overactuated.
  We perform a modal analysis to show that this may be an advantage,
  as the redundant degrees of freedom can be used to shift the natural frequencies by applying tension to the coupling module.
  To demonstrate the validity of our approach, we perform a milling experiment using our coupled system.
  An external measurement system is used to show that tensioning the coupling module causes a deformation of the system.
  We further show that this deformation is static over the tool path and can be compensated for.
\end{abstract}

\begin{keyword}
Milling \sep Dynamics \sep Robotics



\end{keyword}

\end{frontmatter}



\section{Introduction}
\label{introduction}

The field of industrial robots initially began with robots as universal part-handling devices in the automotive industry.
From these humble beginnings, industrial robots have evolved into flexible manufacturing systems used in a variety of industries.
One increasingly popular application is the use of robots for machining tasks.
The main advantages of robots in machining are their flexibility and large workspace.
However, the low dynamic stiffness of serial kinematic chains is a major disadvantage for machining tasks.
The field of industrial robotics has therefore long been concerned with optimizing robotic milling applications.
A good overview of these efforts can be found in the following reviews~\cite{review_1}\cite{review_2}.
The four main challenges according to ~\cite{review_1} are the deformation of the robot,
dealing with the heterogeneous stiffness distribution of the robot, incorporating the complicated robot dynamics, and suppressing chatter.\\
Common approaches thus focused on optimizing the workpiece position such that the robot's stiffness is maximized~\cite{previous_work}\cite{stiffness_placement}.
Other approaches integrate specialized tools to directly counter the excitation forces~\cite{special_tools}.\\
\\
A recent approach proposes enhancing robot stiffness by coupling two robots at the flanges~\cite{edgar}.
In a simplified model, a single robot can be seen as a single mass spring damper system.
The mass is the tool, the spring is the robot's stiffness and the damper is the robot's damping.
In the real system, both stiffness $K$ and damping $D$ are functions of the robot's joint configuration $q$.
In this model coupling two robots at the flanges is akin to adding a second spring and damper in parallel, effectively doubling the system's stiffness.
Previous studies have shown that this modification also leads to a more uniform stiffness distribution across the workspace~\cite{edgar}.
While a higher static stiffness improves surface finish by reducing tool deflection, it doesn't eliminate chatter.
However, the coupling not only doubles stiffness but also adds redundant actuators.
In our simplified model, these additional actuators would enable the shifting of the second spring's resting position
by moving the second robot to tension the coupling module. In the theoretical model, this does not change the resonant frequency but only decreases its amplitude.
We hypothesize that due to the nonlinearities of the real system, the resonant frequency will shift.
In turn, shifting natural frequencies has been shown as an effective strategy to suppress chatter~\cite{chatter_surpression} in robotic applications.
We thus propose a new type of milling system that consists of two industrial robots that are coupled at the flanges.
We perform a modal analysis to show that the natural frequencies of the system can be shifted by applying tension to the coupling module.
We then perform a milling experiment to show that it is possible to still machine a workpiece even under tension.
Since the tension is bound to cause a deformation of the system we also show that it is possible to compensate for this deformation.\\
\\
The rest of this paper is structured as follows.
In Section \ref{related_work} we discuss related work.
In Section \ref{coupled_robot_dynamics} we introduce the proposed robotic milling system.
In Section \ref{modal_analysis} we perform the modal analysis investigating the shift in natural frequencies.
The milling experiment is discussed in Section \ref{milling_experiment}.
We conclude the paper in Section \ref{conclusion} with a discussion of the results and an outlook on future work.

\section{Related work}
\label{related_work}

Historically, the advent of robotic milling in the 1980s marked the beginning of efforts to address these challenges \cite{Terrier2004}. Strategies have ranged from mechanical interventions, such as increasing the stiffness of the robot with passive or active elements, to process and control strategies, including the adaptation of process parameters, the exploitation of kinematic redundancies \cite{Mousavi2018}, the optimization of workpiece positioning \cite{Cvitanic2020}, and the design of robot trajectories informed by stiffness models.

In the quest to optimize robotic milling operations, the development of comprehensive stiffness models stands out as a significant advance. These models address both static and dynamic analyses, facilitating the establishment of performance indices such as stability regions. These indices are essential for determining the most productive milling parameters within the constraints of given robot configurations. Dynamic system analysis and trajectory planning techniques have been key to mitigating deflection and chatter. Approaches have included both offline \cite{Lehmann2012} and online compensation strategies \cite{Lin2023}, using model-based methods \cite{Nguyen2019}, model-free approaches, and posture- or force-based strategies \cite{Wu2021}.

The concept of overactuating milling systems, aiming to exploit the inherent stiffness of parallel kinematic structures, has been explored through various initiatives, including the proposition of an over-actuated parallel kinematic milling system in the early 2000s \cite{2001_redundant_milling}.
Their justification for parallel kinematics is the same as ours, namely that it is inherently stiffer.
However, the redundant degrees of freedom were primarily used to remove singularities from the system.
The same is true for other works such as~\cite{Xie2014} and~\cite{Shin2012}.
The extra degrees of freedom were thus deployed in serial which ultimately just reduces the overall stiffness of the system.
Redundantly actuated milling systems with closed kinematic chains have thus not been studied in depth.\\
\\
An example of such a system is presented in~\cite{xin}, where two KUKA iiwa robots are coupled.
The optimization objective was to minimize workpiece deformation caused by external forces.
Unlike typical milling scenarios, only static stiffness was considered since the applied force in the paper was static.
The redundant degrees of freedom were treated as a problem to be solved instead of a feature to be exploited.
This is similar to other collaborative robotics applications such as~\cite{collaborative_robots} where multiple mobile manipulators must distribute the load of an object they carry.
The optimization of such systems is typically performed at the force level.
This is not possible in realistic milling applications where industrial robots don't offer the necessary joint torque interfaces.
Our approach will thus only control the robot at the position level and use full-scale industrial robots
as well as the coupling module introduced by~\cite{coupling_module}.

\section{The proposed robotic milling system}
\label{coupled_robot_dynamics}
\begin{figure*}[h]
  \centering
  \includegraphics[width=\textwidth]{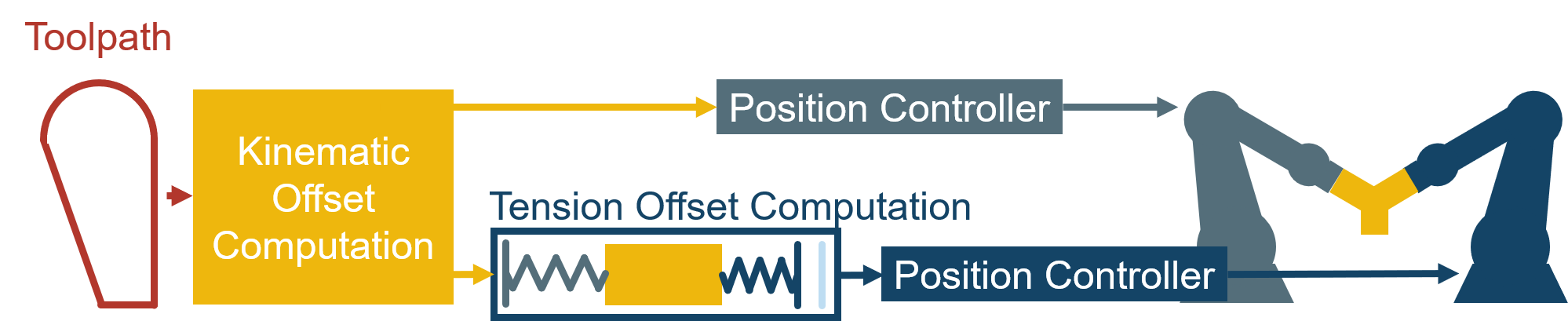}
  \caption{Block diagram of the proposed robotic milling system. The toolpath gets translated for both robots, additionally one robot introduces tension via an offset in position using a linear spring model.}
  \label{fig:system_overview}
\end{figure*}

\begin{figure}[h]
  \centering
  \includegraphics[width=\columnwidth]{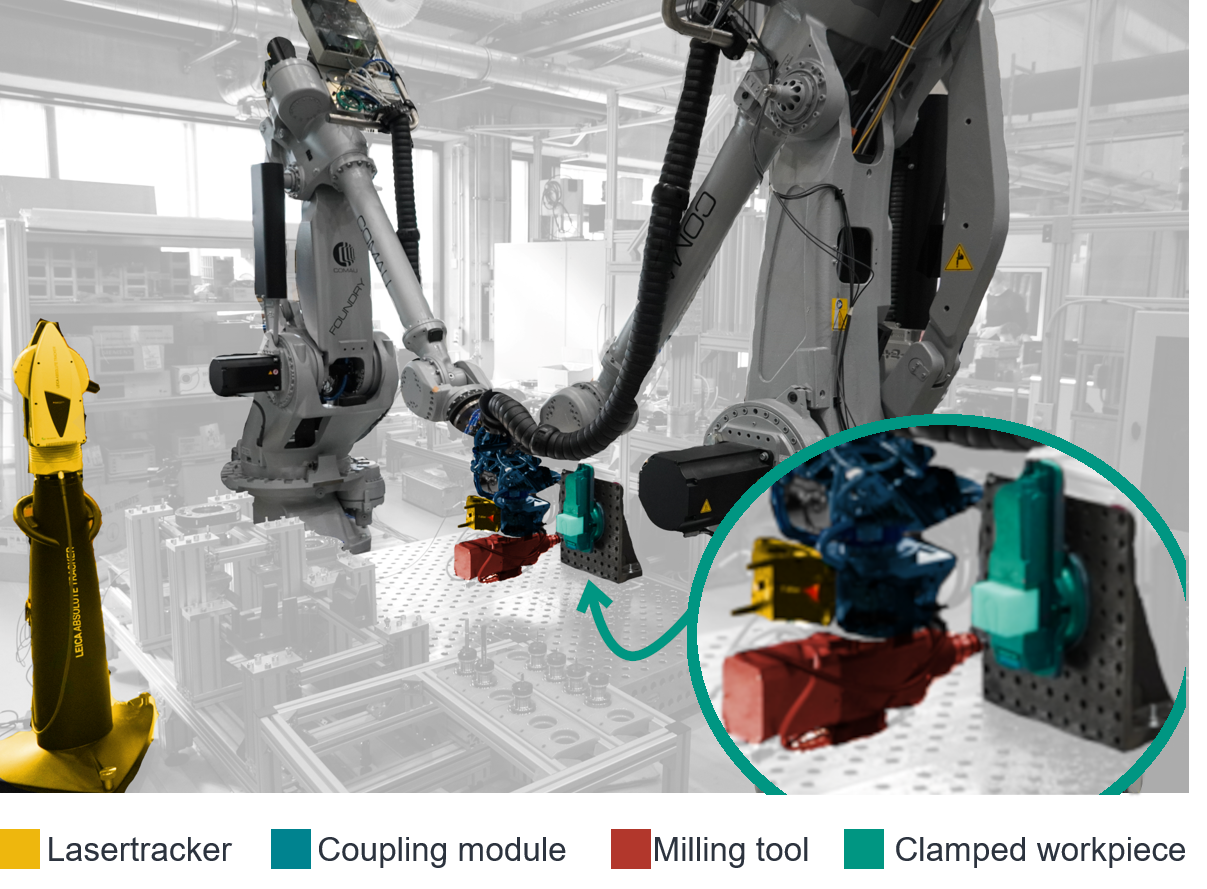}
  \caption{Lab setup of the proposed robotic milling system. Both robots are coupled and a spindle and a 6 DOF Lasertracker target is attached.}
  \label{fig:system_picture}
\end{figure}

The proposed robotic milling system consists of two identical industrial robots as well as a coupling module.
The coupling module is a passive mechanical device that connects the two robots at their flanges.
For safety reasons, it may be equipped with force/torque sensors at both connectors to measure internal forces.
For this paper, the coupling module from~\cite{coupling_module} is used.
Two Comau NJ290 3.0 robots are used as industrial robots. Each robot has a payload of 290 kg and a reach of 3 meters.
The robots are positioned at a distance of 4.25 m from each other, allowing them to comfortably reach a workspace of 1000 mm x 1000 mm x 1000 mm.
While slightly smaller than the reach of a single robot this is still a very large workspace for a milling system.
Even larger workspaces might also be possible but would require the robots to be positioned further apart which in turn reduces the stiffness of the system.
The positioning of the robots is thus a tradeoff between workspace and stiffness.\\
\\
The robots are controlled using a single Siemens Sinumerik One NC controller,
which controls both robots through joint interpolation and path planning, known as Synchronized Motion by Siemens.
This function achieves geometric coupling and setpoint synchronization.
This is important since time delays between the robots could tear the coupling module apart.
Tensioning the system is achieved by offsetting the setpoint of the second robot.
The offset can be calculated from a desired tension force using the stiffness model of the combined system.
This model assumes that the robots are perfectly rigid except for the joints and that the coupling module is a six-dimensional linear spring.
The spring parameters were determined using a finite element analysis.\\
\\
The coupling module is equipped with a quick-change system for the tool.
The tool itself is a water-cooled high-speed spindle with up to 24000 revolutions per minute.
Toolholding is achieved using a hollow shaft taper system.
A schematic overview system can be seen in Figure \ref{fig:system_overview}, while a picture of the system can be seen in Figure \ref{fig:system_picture}.

\section*{Modal analysis}\label{modal_analysis}

\begin{figure}[h]
    \centering
    \includegraphics[width=\columnwidth]{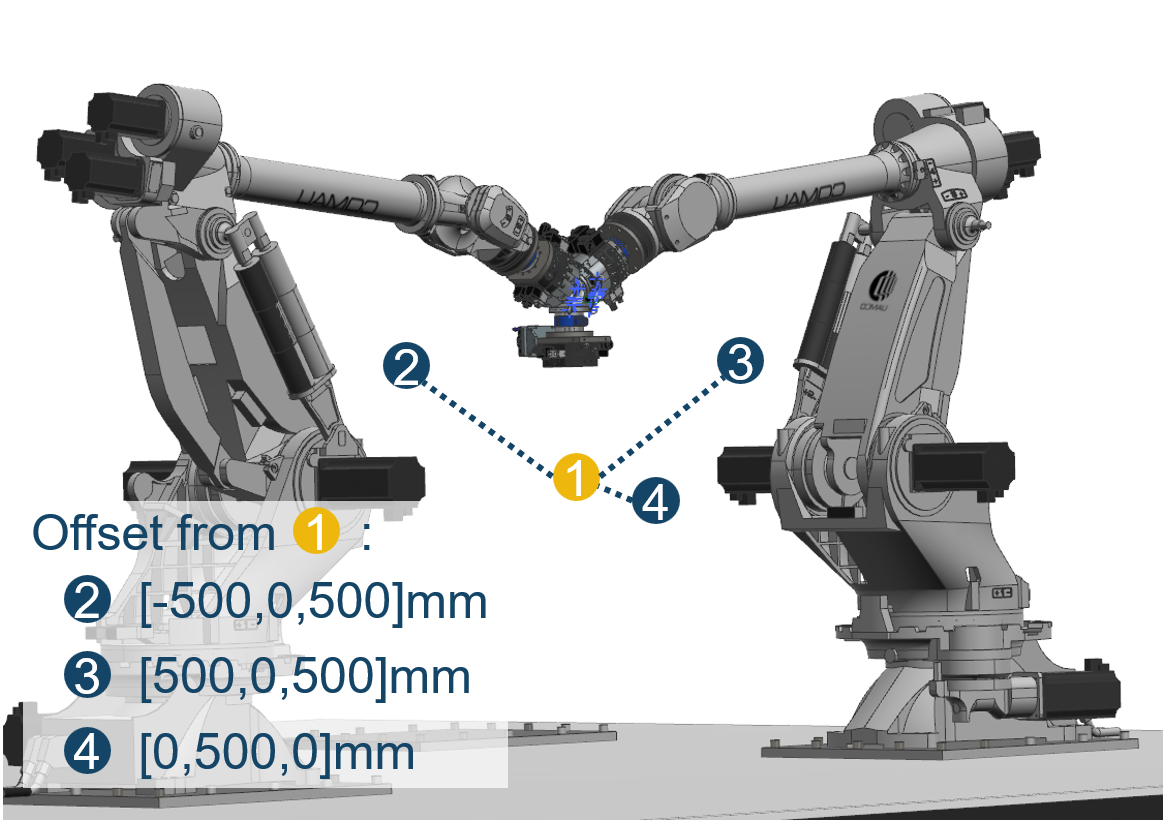}
    \caption{Visualization of the modal analysis positions with relative offsets from the starting position (1). The first dimension points from the left robot to the right.}
    \label{fig:modal-setup}
\end{figure}

Modal impact experiments were performed to test the hypothesis of the frequency shift due to different levels of tension forces induced by the robots.
The experiments were performed on the coupled robot system, with impact hammer tests and a 3-axis accelerometer.
Impact tests were performed at 4 points in the workspace (see Fig.\ref{fig:modal-setup}). At each point impacts in all 3 major directions were performed.\\
\\
The measurement positions in the task space were defined to get an overview of the frequency response in the main working area.
This was achieved by first defining a center position (P1) which is in the middle of the working area.
Positions P2, P3, and P4 were then defined relative to this center position to test the response towards the edges of the working area.
Points P2 and P3 were shifted in the x-y direction to test the symmetry of the system, while position P4 was shifted in the y direction to test the response perpendicular to the robot arms.
An overview of all positions can be seen in Figure \ref{fig:modal-setup}.
Each point and direction was tested using increasing tension forces of 0N, 500N, 1400N, and 2000N.
However, since the tension model is not yet perfect, the final tension slightly deviated in each experiment.
The actual tension forces were measured using the force/torque sensors in the coupling module.\\
\\
The 3-DOF acceleration sensor was attached using wax to a quick change module at the bottom of the coupling module. This made it possible to keep the sensor position unchanged between the different measurement positions.
The impact was aimed at predefined surfaces near the sensor and was delivered using an impact hammer with a force sensor to measure the impact force.

\begin{figure}[h]
    \includegraphics[width=\columnwidth]{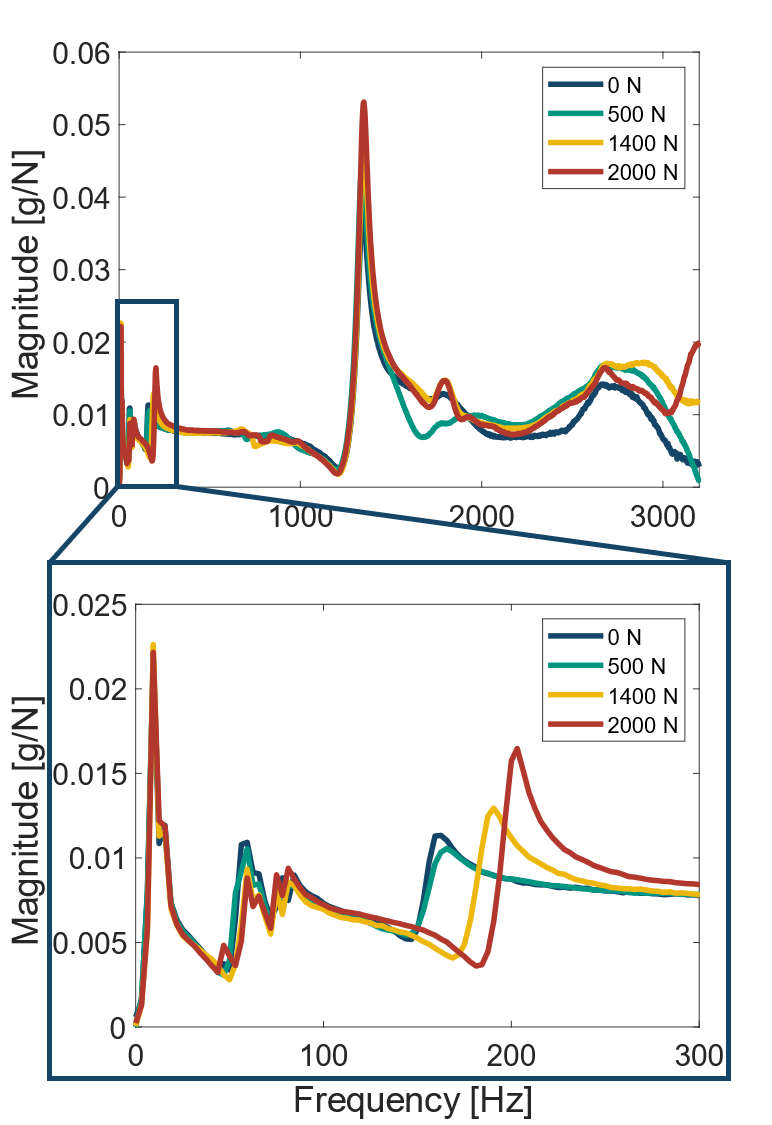}
    \caption{Frequency response for first measure position and impact in positive x direction over tension forces. The peaks in compliance shift towards higher frequencies with an increase of tension force.}
    \label{fig:modal-res-p1}
\end{figure}
\begin{figure}[h]
    \centering
    \includegraphics[width=\columnwidth]{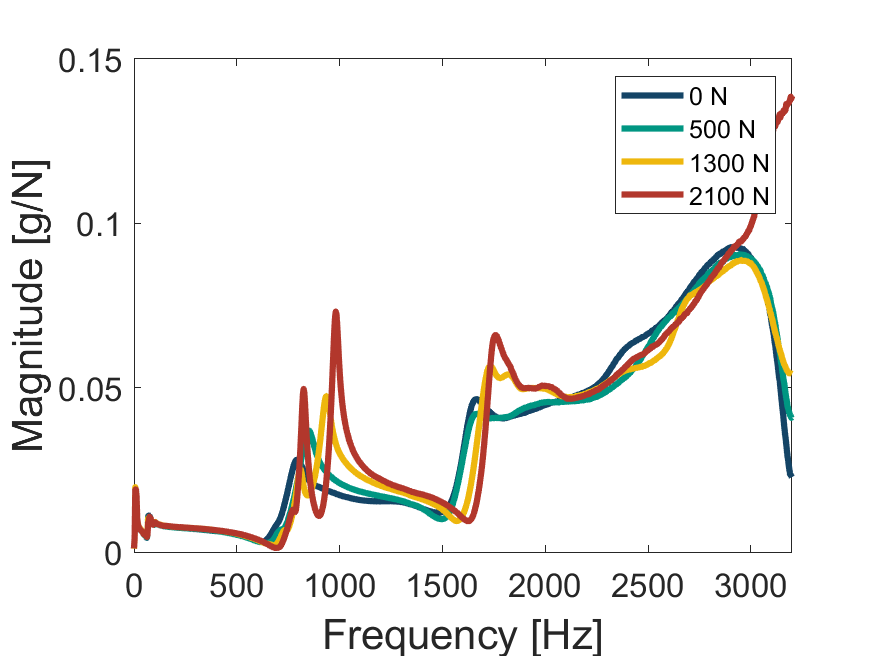}
    \caption{Frequency response for first measure position and impact in +Y over tension forces. No shifts in frequency under 500hz were measured, and shifting starts at 700hz for increasing tension forces.}
    \label{fig:modal-res-p1-y}
\end{figure}
\begin{figure}[h]
    \centering
    \includegraphics[width=\columnwidth]{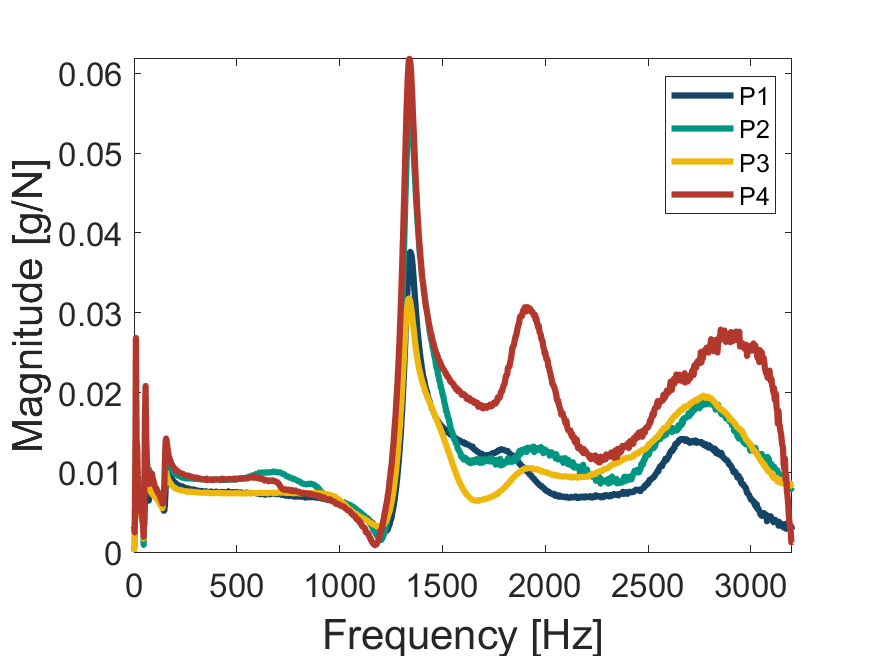}
    \caption{Frequency response for all measure positions in +X. Very similar behaviour of frequency responses at different positions.}
    \label{fig:all_positions}
\end{figure}
\begin{figure}[h]
    \centering
    \includegraphics[width=\columnwidth]{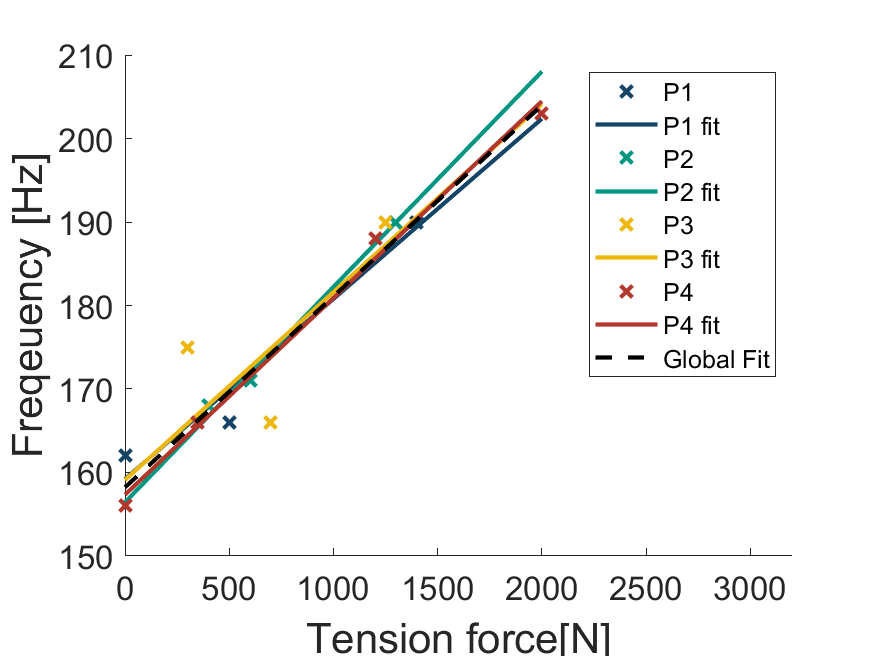}
    \caption{Shift in natural frequency over tension forces with individual as well as global linear fits. Using a global linear approximation of the frequency shifts is feasible.}
    \label{fig:shift_vs_tension}
\end{figure}

The resulting frequency response of measurement point 1 in x direction can be seen in Figure \ref{fig:modal-res-p1} for different levels of tension forces.
Here we can identify one main region of interest, seen in the lower plot.
It shows a frequency shift of the natural frequency in x direction at 159 Hz which can be shifted to 165 Hz with 500N of tension forces, 190 Hz at 1400N, and 202 Hz at 2000N of tension forces.
This proves our hypothesis that tension forces can be used to shift the natural frequencies of the system.
However, it is notable that we don't shift the lower natural frequencies which would also be very relevant for chatter suppression.
One possible explanation for this is that for lower frequencies the single-mass spring damper model seems to hold better.
This would also predict a decrease in amplitude for higher tension which can indeed be observed for the lower natural frequencies.\\
\\
Comparing the results of the x direction with the results of the y direction which is plotted in Figure \ref{fig:modal-res-p1-y} we observe that the frequency shift starts at much higher frequencies.
While we observe a similar shift in natural frequencies, the natural frequency is too high to be relevant for chatter suppression.
The reason for this discrepancy is that the x direction is in line with the robot arms while the y direction is perpendicular to them.
This means that the tension forces are not as effective in the y direction as they are in the x direction.
Future work could investigate if a coupling at right angles can decrease this directional dependency.\\ 
\\
Comparing the different measurement positions under no tension force, as plotted in Figure \ref{fig:all_positions}, we observe that the natural frequencies are very similar.
Again we also observe the largest deviation at P4, which is likely due to it being perpendicular to the robot arms.
Still, the similarity might be surprising as each position corresponds to a different joint configuration which should result in a different stiffness of the system.
However, these results just confirm the work of~\cite{edgar} that coupling the robots leads to a more uniform stiffness distribution across the workspace.
Indeed, the results indicate that this is also true for the dynamic stiffness.
A more uniform stiffness distribution is thus another major advantage of the proposed system as it makes the natural frequencies easier to predict even without the application of tension forces.
\\
If we additionally want to use the tension forces to suppress chatter, we need to be able to predict the shift in natural frequencies.
For this reason, we plotted the shift in natural frequency over the tension forces in Figure \ref{fig:shift_vs_tension}.
As we can see the shift of the natural frequencies seems to roughly follow a linear relationship with the tension forces.
Furthermore, the shift in natural frequencies seems to be very similar for all measure positions.
This indicates that it is possible to predict the shift in natural frequencies for different measure positions using a global linear fit.
This would make it possible to shift natural frequencies in a controlled manner to suppress chatter in a milling process.
However, this still requires an accurate system of predicting internal tension forces based on robot positions.
Here a more accurate model is needed to predict the tension forces based on the robot positions.
This is however left for future work.

\section{Milling experiment}
\label{milling_experiment}
\begin{figure}[h]
    \centering
    \includegraphics[width=\columnwidth]{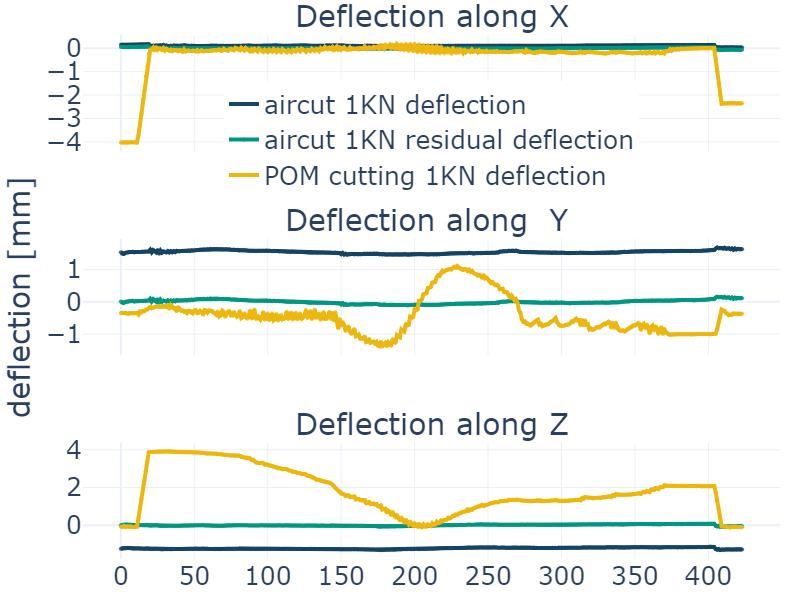}
    \caption{Deformation of the coupling module under 1000N of tension force relative to an air cut with no tension. Deformation is mostly constant and therefore easy to compensate for, as shown by the residual deflection line.}
    \label{fig:deformation}
\end{figure}

While the modal analysis has shown that chatter suppression is possible using tension forces, we have yet to prove that it is possible to still machine a workpiece under tension forces.
For this reason, a milling experiment was performed using the proposed robotic milling system.
The target path can be seen in Figure \ref{fig:system_overview} and consists of two linear and one circular subpath.
The workpiece is a 100mm x 80mm x 80mm cube made of polyoxymethylene (POM).
The reason for using POM is that it is a technical plastic that should not apply any significant force to the system.
This is important since we foresee that the tension forces will cause a deformation of the system.
To properly measure this deformation we need to prevent significant process forces which would also cause deformations.
The measurement of the coupling module position was performed using a Leica AT960MR laser tracker with a base measurement accuracy of 15$\mu m$.
The laser tracker was positioned on the opposite side of the workpiece to the coupling module.
The workpiece itself was vertically clamped to the table as seen in Figure \ref{fig:system_picture}.
To measure the effects of tension forces on the deformation as well as prove that milling is still possible under tension forces, three experiments were performed.
The first experiment is an air cut where the tool is moved along the target path without any tension.
This serves as a reference for the following experiments.
In a second experiment, the air cut was repeated under 1000N of tension force.
Finally, the third experiment was a real cut under 1000N of tension force.
To investigate the deformation of the system the difference between the tensioned experiments to the reference experiment was plotted in Figure \ref{fig:deformation}.
Looking at the air cut under 1000N of tension force we can see a deformation of the coupling module of up to 2 mm in the y direction and 1.2 mm in the z direction.
Fortunately, this deformation seems to be constant over the whole path.
This makes it possible to compensate for the deformation by calculating the rotation and translation between the untentioned and tensioned experiments.
Applying this transformation to our measured tensioned path, we see that the deformation is almost completely compensated.
This is a promising result as without any compensation the milling system would be unusable under tension forces.
Future work should therefore investigate how the deformation can be predicted based on the tension forces and robot positions.\\
Even though we are using a technical plastic for the workpiece, we still see process force-related deformation in the POM cutting experiment.
However, this is likely not a function of the compliance of the coupling module itself but the contribution of the rubber pads currently used for safety reasons against the tension system.
Future work is already underway for a more comprehensive force controller that no longer requires these rubber pads.

\section{Conclusion}
\label{conclusion}
In this paper, we have proposed a new robotic milling system composed of two industrial robots coupled at the flanges.
We postulated that the redundant degrees of freedom can be used to shift the natural frequencies of the system by applying tension to the coupling module.
The modal analysis experiments confirmed this hypothesis and further showed that the natural frequencies are very similar for different measurement positions.
Thus highlighting another advantage of coupled robotic milling.
We also showed that the relationship between tension forces and natural frequencies is roughly linear and similar for different measure positions.
Nevertheless, the experiments showed areas of improvement relating to the design of the coupling module and the prediction of tension forces.
Both will be the focus of future work.
Finally, we showed that it is possible to still machine a workpiece under tension forces and that the deformation of the system can be compensated for.
However, for a more comprehensive investigation, these experiments should be repeated for different materials and tension forces.
Here a more thorough model is needed to predict the deformation of the system based on the tension forces and robot positions.
We nevertheless believe that the work presented in this paper shows that milling using two mechatronically coupled robots is a promising approach for chatter suppression in robotic milling applications.

\section{Acknowledgement}
This research is funded by the Europäische Fonds für regionale Entwicklung (EFRE, Ministry of Science, Research and Arts) - KoppRob(n).
The authors would like to thank the Ministry of Science, Research and Arts of the Federal State of Baden-Wuerttemberg for the funding of the lasertracker within the framework of the "Future Mobility Grants" of the InnovationCampus Future Mobility (ICM).

\bibliographystyle{elsarticle-num}
\bibliography{bibliography}

\end{document}